# Towards Explainable and Language-Agnostic LLMs: Symbolic Reverse Engineering of Language at Scale

Walid S. Saba

*Institute for Experiential AI at Northeastern University*
*Portland, ME USA*
*w.saba@northeastern.edu*

## Abstract

Large language models (LLMs) have achieved a milestone that undeniably changed many held beliefs in artificial intelligence (AI). However, there remains many limitations of these LLMs when it comes to true language understanding, limitations that are a byproduct of the underlying architecture of deep neural networks. Moreover, and due to their subsymbolic nature, whatever knowledge these models acquire about how language works will always be buried in billions of microfeatures (weights), none of which is meaningful on its own, making such models hopelessly unexplainable. To address these limitations, we suggest combining the strength of symbolic representations with what we believe to be the key to the success of LLMs, namely a successful bottom-up reverse engineering of language at scale. As such we argue for a bottom-up reverse engineering of language in a symbol-ic setting. Hints on what this project amounts to have been suggested by several authors, and we discuss in some detail here how this project could be accomplished.

## 1   Introduction

In general, scientific explanation proceeds in one of two directions: by following a top-down strategy or by following a bottom-up strategy (Salmon, 1989). For a top-down strategy to work, however, one must have access to a set of *general principles* to start with and this is certainly not the case when it comes to thought and how our minds externalize our thoughts in language. Nevertheless, decades of work in natural language processing (NLP) marched on inspired by generative linguistics, where an innate language faculty and a Universal Grammar were postulated (Chomsky, 1956), cognitive linguistics, where it was postulated that we metaphorically build our linguistic apparatus on top of a set of idealized cognitive models (ICMs) (Lakoff, 1987), or model-theoretic semantics (Montague, 1974), where it was postulated that natural languages, like formal languages, can be precisely specified using the tools of mathematical logic. However, in all cases there was very little in terms of established knowledge that these theories started from. In retrospect, then, and lacking any general principles one can speak of about our language (and *the language of thought*) it is no surprise that the bottom-up method succeeded where decades of top-down work in NLP failed to deliver. Moreover, and due to the intricate relationship between language and knowledge, this is perhaps the reason why much work in knowledge representation and ontology also failed (Sowa, 1995; Lenat, 1990), since most of this work amounted to pushing, in a top-down manner, various metaphysical theories of how the world is supposedly structured and represented in our minds, and again without any established general principles to start from.

On the other hand, a little more than a decade of work in bottom-up reverse engineering of language has produced very impressive results. With the release of GPT-4 it has become apparent that large language models (LLMs), that are essentially a massive experiment in a bottom-up reverse engineering of language, have crossed some threshold of scale at which point there was an obvious qualitative improvement in their capabilities[1]. It is our opinion that these capabilities mark a milestone, and not just a computational one, but a theoretical one, and we think it is one that linguists, psychologists, philosophers, and cognitive scientists must reflect on. In particular, we believe that a number of reservations expressed by luminaries in the philosophy of language and the philosophy of mind concerning the possibility of machine understanding are now questionable, if not outright irrelevant. For example, we believe the arguments of

---

[1] GPT stands for 'Generative Pre-trained Transformer', an architecture that OpenAI built on top of the transformer architecture introduced in (Vaswani, et. al., 2017).

Hubertus Dreyfus (1972) who suggested that computers will never know what is *relevant* in a given situation, are not very convincing anymore since GPT-4 certainly replies with 'relevant' content in response to some prompt. Moreover, we believe the thought experiment devised by the philosopher John Searle (1980), one that questioned the possibility of machines exhibiting any semantics, to also be somewhat irrelevant now. While lots of ink has been spilled on what has become known by the Chinese Room Argument (CRA), current capabilities of LLMs clearly demonstrate not only a mastery of syntax but quite a bit of semantics too. Indeed, what the massive experiments that lead to LLMs have shown is that quite a bit of semantics, and even quite a bit of commonsense knowledge, both of which are clearly encoded in our everyday linguistic communication, can be uncovered in a bottom-up reverse engineering process[2]. But, in our opinion, this is where the good news ends for LLMs.

## 2  Limitations of LLMs

To begin with, and despite their relative success, we should remain cognizant of the fact that LLMs models are not (*really*) 'models of language' but are statistical models of the regularities found in linguistic communication. Models and theories should explain a phenomenon (e.g., $F = ma$) but LLMs are not *explainable* because explainability requires structured semantics and reversible compositionality that these models do not admit (Saba, 2023). As shown in figure 1 below, the output of a neuron that is obtained by applying some activation function on the output of some weighted sum is not invertible. In figure 1*a* we use here 12 (a scalar tensor) as an example but this applies to a tensor/vector of any dimension. That is, vectors decomposition is undecidable because the composite function that performs linear combination of addition and multiplication is not invertible. On the other hand, composition in symbolic systems is invertible as shown in figure 1*b*, since in symbolic systems we maintain a semantic map of the computation (e.g., an abstract syntax tree). The point here is that due to their subsymbolic nature, whatever 'knowledge' LLMs acquire about language will always be buried in billions of microfeatures (weights), none of which is meaningful on its own. In addition to the lack of explainability, LLMs will always generate biased and toxic language since they are susceptible to the biases and toxicity in their training data (Bender et. al., 2021). Moreover, and due to their statistical nature, these systems will never be trusted to decide on the "truthfulness" of the content they generate (Borji, 2023)[3]. Note that none of these problematic issues are a function of scale but are paradigmatic issues that are a byproduct of the architecture of deep neural networks (DNNs).

---

[2] While this is not our immediate concern, but we believe this is what John Searle missed in his CRA thought experiment, namely that syntax and semantics are two sides of the same coin, and that mastering syntax implicitly means mastering quite a bit of the semantics that is embedded in the syntax, as has clearly been demonstrated by LLMs. It is for this reason that we can make syntactically valid expressions that are meaningless, but we cannot have a meaningful expression if it was not syntactically valid!

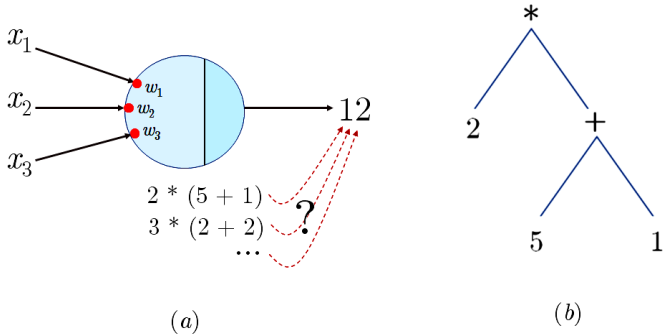

**Figure 1**. Compositional computation in subsymbolic systems (e.g., DNNs) is not invertible. The decomposition of 12 into its original components is undecidable (*a*), unlike symbolic systems (*b*), where there are structures that 'save' the semantic map of the computation.

There are other limitations that are also not a function of scale, but a byproduct of the underlying architecture. LLMs are based on the architecture of DNNs that do not admit any symbolic representations and are thus purely extensional models and would therefore fail to make the correct inferences in intensional contexts. Consider the example shown in figure 2(a) below. While 'Aristotle' and 'the tutor of Alexander the Great' have the same extension (they both refer to the same object), as objects of cognition they have different intensions (senses), thus their extensional equality should not license a replacement of one for the other as GPT-4 does, resulting in the absurd conclusion that '*perhaps the tutor of Alexander the Great was not the tutor of Alexander the Great*'. In figure 2(*b*) we have a similar situation were replacing Paris with '*the most populous city in France*' – while extensionally valid, also results in a non-truth, since Mary's stating her desire to visit Paris does not entail Mary's stating her desire to visit the most populous city in France. In addition to failing in intensional contexts LLMs cannot be relied upon in contexts where the resolution of scope ambiguities requires access to subtle commonsense knowledge. For example, in figure 2(c) GPT-4 does not correctly interpret "two museums" as "many museums" since, from the standpoint of commonsense, the correct reading should be "every tourist is taken to two museums by some student"[4]. Note that these tests are not exotic or farfetched and can always be reproduced as they are examples that are, in theory, beyond the capabilities of purely extensional LLMs (note that due to some added randomness, to simulate 'creativity', the same prompts may produce slightly different results).

---

[3] Truth is not approximate, and not only when it comes to mathematical facts. Much like it is meaningless to speak of the probability of (6 = 2 * 4), it is also meaningless to assign any probability to the result of the database query '*Is John Smith the sales manager in our Chicago branch?*' – facts either are, or they are not.

There are other limitations that are also not a function of scale, but a byproduct of the underlying architecture. LLMs are based on the architecture of DNNs that do not admit any symbolic representations and are thus purely extensional models and would therefore fail to make the correct inferences in intensional contexts. Consider the example shown in figure 2(a) below. While 'Aristotle' and 'the tutor of Alexander the Great' have the same extension (they both refer to the same object), as objects of cognition they have different intensions (senses), thus their extensional equality should not license a replacement of one for the other as GPT-4 does, resulting in the absurd conclusion that '*perhaps the tutor of Alexander the Great was not the tutor of Alexander the Great*'. In figure 2(*b*) we have a similar situation were replacing Paris with '*the most populous city in France*' – while extensionally valid, also results in a non-truth, since Mary's stating her desire to visit Paris does not entail Mary's stating her desire to visit the most populous city in France. In addition to failing in intensional contexts LLMs cannot be relied upon in contexts where the resolution of scope ambiguities requires access to subtle commonsense knowledge. For example, in figure 2(c) GPT-4 does not correctly interpret "two museums" as "many museums" since, from the standpoint of commonsense, the correct reading should be "every tourist is taken to two museums by some student"[5]. Note that these tests are not exotic or farfetched and can always be reproduced as they are examples that are, in theory, beyond the capabilities of purely extensional LLMs (note that due to some added randomness, to simulate 'creativity', the same prompts may produce slightly different results).

---

[4]A list of such examples involving intensional contexts as well as examples that involve commonsense reasoning (e.g. in the context of quantifier scope) can be found here `https://medium.com/ontologik/a-serious-chat-with-chatgpt-99e7de8d68c2`.

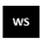

**Figure 2**. Examples showing how LLMs fail in modal and intensional contexts.

---

[5]A list of such examples involving intensional contexts as well as examples that involve commonsense reasoning (e.g. in the context of quantifier scope) can be found here `https://medium.com/ontologik/a-serious-chat-with-chatgpt-99e7de8d68c2`.

So where do stand now? From our discussion thus far it would seem that the glass is half full and half empty. On one hand, LLMs have clearly proven that one can get a handle on syntax and quite a bit of semantics in a bottom-up reverse engineering of language at scale; yet on the other hand what we have now are unexplainable models that do not shed any light on how language actually works and how we externalize our thoughts in language. Since we believe that the relative success of LLMs is not a reflection on the symbolic vs. subsymbolic debate but is a reflection on (appropriately) adopting a bottom-up reverse engineering strategy, we think that combining the advantages of symbolic representations with a bottom-up reverse engineering strategy is a worthwhile effort. The idea that word meaning can be extracted from how words are actually used in language is not exclusive to linguistic work in the empirical tradition, but in fact it can be traced back to Frege, although there were more recent philosophical and even computational proposals on what this project amounts to. Below we discuss these proposals in some detail.

## 3     Concerning "The Company a Word Keeps"

The genesis of modern LLMs is the *distributional semantics hypothesis* which states that the more semantically similar words are, the more they tend to occur in similar contexts – or, similarity in meaning is similarity in linguistic distribution (Harris, 1954). This is usually summarized by a saying that is attributed to the British linguist John R. Firth that "you shall know a word by the company it keeps". When processing a large corpus, this idea can be used by analyzing co-occurrences and contexts of use to approximate word meanings by word embeddings (vectors), that are essentially points in multidimensional space. Note, however, that this part of the story covers only what is called *lexical semantics*, which is the study concerned with word meanings. In particular, this part of the story does not address modeling syntactic rules nor compositional semantics, by which the meaning of larger linguistic units is obtained as some function of the meaning of the parts and how they appear together. Instead, the meaning of larger linguistic units in this tradition was usually obtained by some weighted vector addition operation, although there were many attempts to combine traditional compositional semantics with vector semantics in what has come to be known by compositional distributional semantics (CDS). See (Baroni, et. al., 2014) for an excellent review of this work.

While word embeddings can approximate lexical semantics (word meanings), it was not until the transformer model (Vaswani, et. al., 2017) that embeddings started the encoding of syntax. That is, what transformers and multiple attention heads did is create embeddings for 'valid' sequences and not just words. But how many of these sequences can one encode? Apparently, it has taken a massive network with over 500 billion encodings to master the syntax of language. In this regard, it is worth mentioning here an astute observation made by Stephen Wolfram (2023) regarding the size of these deep networks, namely that "the size of the network that seems to work well is so comparable to the size of the training data". In other words, it would seem that (roughly) an additional parameter (weight) was required for every additional token in the corpus. If this correlation is not accidental then it is another indication that such models cannot provide an explainable model/theory for how language works since it would mean that what these models are doing, in effect, is encoding (memorizing) all possible combinations of how words may appear in any sequence of words, which is hardly a theory of linguistic communication.

In summary, transformers with attention, along with massive scale, have allowed for a qualitative leap in the linguistic capabilities of LLMs. Still, at the root of this bottom-up reverse engineering of language is the concept of 'the company a word keeps' and the distributional semantics hypothesis that, unlike top-down approaches, "reverse engineers the process and induces semantic representations from contexts of use" (Boleda, 2020). But nothing precludes this ingenious idea from being carried out in a *symbolic* setting. In other words, the 'company a word keeps' can be measured in several ways, some of which, incidentally, have been discussed since Frege. We turn to this subject next.

## 4   Symbolic Reverse Engineering of Language

In discussing possible models (or theories) of the world that can be employed in computational linguistics Jerry Hobbs (1985) once suggested that there are two alternatives: on one extreme we could attempt building a "correct" theory that would entail a full description of the world, something that would involve physics and all the sciences; on the other hand, we could have a promiscuous model of the world that is isomorphic to the way we talk it about in natural language. Clearly, what Hobbs is suggesting here is a reverse engineering of language itself to discover how we actually use language to talk about the world we live in. In essence, this is not much different from Frege's Context Principal that suggests to "never ask for the meaning of words in isolation" (Dummett, 1981) but that a word gets its meanings from analyzing all the contexts in which the word can appear (Milne, 1986).

Again, what this suggests is that the meaning of words is embedded (to use a modern terminology) in all the ways we use these words in how we talk about the world. While Hobbs' and Frege's observations might be a bit vague, the proposal put forth by Fred Sommers (1963) was very specific. Again, Sommers suggests that "to know the meaning of a word is to know how to formulate some sentences containing the word" and this would lead, like in Frege's case to the conclusion that a complete knowledge of some word *w* would be all the ways *w* is used and in every possible sentence. For Sommers, the process of understanding the meaning of some word *w* starts by analyzing all the properties *P* that can sensibly be said of w. Thus, for example, [*delicious Thursday*] is not sensible while [*delicious apple*] is. Moreover, and since [*delicious cake*] is also sensible, there must be a common type (perhaps **food**?) that subsumes both *apple* and *cake*. This idea seems similar to the idea of type checking in programming languages. For example, the types in an expression such as '$x + 3$' will only unify (or the expression will only 'make sense') if/when $x$ is an object of type number (as opposed to a tuple, for example). As it was suggested in Saba (2007), this type of analysis can be used not only to discover the dimensions of word meanings, but to 'discover' the ontology that seems to be implicit in all natural languages.

Let us now consider the following naïve procedure for some initial reverse engineering of language, a procedure that was initially suggested in Saba (2007):

1. Consider concepts $C = \{c_1, ..., c_m\}$ and properties $P = \{p_1, ..., p_n\}$.
2. Assume a predicate **app**($p, c$) that is true iff the property $p$ applies to (or is sensible to say of) objects of type $c$, where $c \in C$ and $p \in P$.
3. A set $Cp = \{c \mid \mathbf{app}(p, c)\}$ is generated for all $c \in C$ and all property $p \in P$ such that the property $p$ is applicable to (or can sensibly be said of) $c$.

4.  A concept hierarchy is then systematically discovered by analyzing the subset relationship between the various sets generated.

Applying the above procedure on a fragment of natural language and taking, initially, *C* to be a set of nouns and *P* a set of adjectives and relations that can sensibly be applied to (or can be said of) nouns in *C*, would result in something like the following:

$R_1$ :  **app**(OLD, entity)
$R_2$ :  **app**(HEAVY, physical)
$R_3$ :  **app**(HUNGRY, living)
$R_4$ :  **app**(ARTICULATE, human)
$R_5$ :  **app**(MAKE(human, artifact))
$R_6$ :  **app**(MANUFACTURE(human, instrument))
$R_7$ :  **app**(RIDE(human, vehicle))
$R_8$ :  **app**(DRIVE(human, car))

What the above say, respectively, is the following:
$R_1$  →  in ordinary language we can say OLD of any entity
$R_2$  →  we say HEAVY of objects that are of type physical
$R_3$  →  HUNGRY is said of objects that are of type living
$R_4$  →  ARTICULATE is said of objects that are of type human
$R_5$  →  MAKE holds between a human and an artifact
$R_6$  →  MANUFACTURE relates a human and an instrument
$R_7$  →  RIDE holds between a human and a vehicle
$R_8$  →  DRIVE holds between a human and a car

Note that the above 'findings' would eventually result in a well-defined hierarchy. For example, since a bottom-up reverse engineering of language will ultimately produce **app**(HEAVY, car) and **app**(OLD, car) – that is, since by analyzing our linguistic communication, we would also discover that it is sensible to say 'heavy car' and 'old car' it would seem that car must be a subtype of physical which in turn must be a subtype of entity. Similarly, since it makes sense to say MAKE of everything, we can say we MANUFACTURE, an instrument must be a subtype of an artifact. The fragment of the hierarchy that is implicit in $R_1$ through $R_8$ is shown in figure 3 below.

Note, also, since **app**(ARTICULATE, human) says that 'articulate' is a property that can be said of objects of type human, we can rewrite this fact as **hasProp**(articulation, human), where ARTICULATE is reified (nominalized) as the trope articulation which is an abstract object of type property (see Moltmann, 2013 for more details on such abstract objects). Using the primitive and linguistically agnostic relation **hasProp** what we now have is a relation between two entities, the property of articulation and a human, which effectively states that articulation is a property that is usually attributed to (or said of) objects that are of type human. The same can be done with $R_3$, **app**(HUNGRY, living), resulting in **inState**(hunger, living) to say that any living entity can be in a state of hunger. The result of this discovery process (that produces linguistic knowledge such as $R_1$ through $R_8$) coupled with the nominalization process and using only primitive relations between entities will be no less than discovering (as opposed to inventing) the ontology that seems to be implicit in ordinary language.

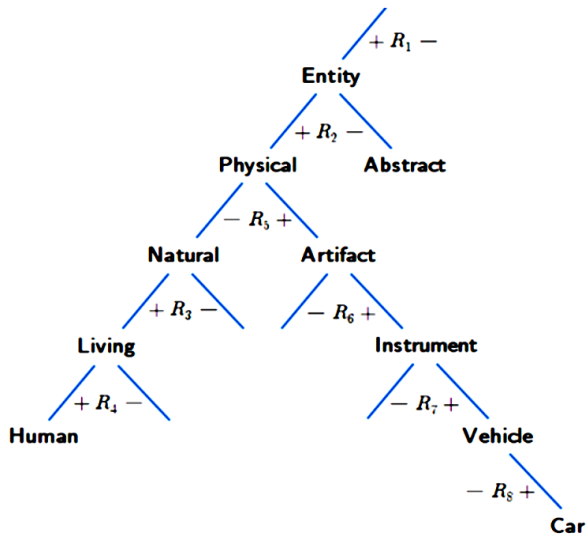

**Figure 3.** The hierarchy that is implicit in the 'discoveries' $R_1$ through $R_8$ above.

Before we discuss the nature of the ontology that seems to be implicit in language use, we need to answer the question of where do these primitive relations come from? That is, how do we discover all these primitive and linguistically agnostic relations, such as **hasProp** and **inState**? The answer to this question lies in the copular ('is' or verb to be). In general, when describing an object or an entity *x* by some property *P* we are, indirectly, making a statement such as '*x* is *P*'. If we analyze the various ways these descriptions can be made, it will lead us to different types of primitive relations, as shown in table 1 below. For example, in saying *Mary is wise*, we are essentially saying that *Mary has the property of wisdom*. Similarly, in saying *Carlos is ill*, we are essentially saying that *Carlos is in the* (physiological) *state of illness*. Analyzing different ways of describing different 'types' of entities would lead us to discover all the language agnostic primitive relations that are summarized in table 2 below.

Here's a summary of the overall process: (*i*) by analyzing a large corpus we can discover all pairs of *c* and *p* for which **app**(*p*, *c*) holds (e.g., **app**(ARTICULATE, human), **app**(HUNGRY, living)); (*ii*) via a nominalization process convert **app**(*p*, *c*) to two entities related by some primitive relation (e.g., **hasProp**(human, articulation), **inState** (human, illness)); (*iii*) construct the ontology implicit in all the discovered relations.

**Table 1.** Discovering the language agnostic primitive relations.

| LINGUISTIC CONTEXT | IMPLICIT PRIMITIVE RELATION |
|---|---|
| Frido *is* a dog | Frido *instanceOf* dog |
| Billy the Kid *is* William H. Boney | Billy the Kid *eq* William H. Boney |
| JFK *is* John Fitzgerald Kennedy | JFK *eq* John Fitzgerald Kennedy |
| Mary *is* wise | Mary *hasProp* wisdom |
| Julie *is* articulate | Julie *hasProp* articulation |
| Jim *is* sad | Jim *inState* sadness |
| Carlos *is* ill | Carlos *inState* illness |
| Sara *is* running | Sara *agentOf* running |
| Olga *is* dancing | John *agentOf* dancing |
| Sara *is* greeted | Sara *objectOf* greeting |
| Sara *is* acknowledged | Sara *objectOf* acknowledgment |
| John *is* 5'10'' tall | John's height *hasValue* 5'10'' |
| Dan *is* 69 years old | Dan's age *hasValue* 69 yrs |
| Sheba *is* running | Sheba *participantIn* running (event) |
| Olga *is* dancing | Olga *agentOf* dancing (activity) |

Table 2. A summary of the language-agnostic primitive relations.

| PRIMITIVE RELATIONS | DESCRIPTION |
| --- | --- |
| $\mathbf{Eq}(x, y)$ | **individual** $x$ is identical to individual $y$ |
| $\mathbf{Part}(x, y)$ | individual $x$ is part of individual $y$ |
| $\mathbf{Inst}(x, y)$ | individual $x$ instantiates **universal** $y$ |
| $\mathbf{Inhere}(x, y)$ | individual $x$ inheres in individual $y$ |
| $\mathbf{Exemp}(x, y)$ | individual $x$ exemplifies property $y$ |
| $\mathbf{Dep}(x, y)$ | individual $x$ depends for its existence on individual $y$ |
| $\mathbf{IsA}(x, y)$ | universal $x$ is a sub-kind of universal $y$ |
| $\mathbf{Precedes}(x, y)$ | individual process $x$ precedes individual process $y$ |
| $\mathbf{HasParticipant}(x, y)$ | individual $y$ participates in individual occurrent $x$ |
| $\mathbf{HasAgent}(x, y)$ | individual $y$ is agent of individual occurrent $x$ |
| $\mathbf{Realizes}(x, y)$ | individual process $x$ realizes individual function $y$ |
| $\mathbf{TypeOf}(x, \mathbf{t}) = (x :: \mathbf{t})$ | individual $x$ is an object of type $\mathbf{t}$ |

## 5 Dimensions of Word Meanings

What we have suggested thus far is a bottom-up reverse engineering of language using the predicate **app**(*p*, *c*) that effectively generates sets for all nouns *c* that the property *p* is applicable of. This in turn can be converted into a triple **R**(entity, entity) corresponding to ([entity] → (**R**) → [entity]) after all the concepts have been reified, where **R** is a primitive and language agnostic relation. Since every entity can now be defined by primitive relations, these primitive relations would now represent the dimensions of word meanings. In figure 3 we show these dimensions for (one of) the meanings of the word book.

As shown in figure 3, one meaning of the word *book* (namely "a written work or composition that has been published") is an entity (*i*) that can be the agent of a changing event (as in '*Das Kapital changed many opinions on capitalism*'); (*ii*) that can have the popularity property (as in '*The Prince is a popular book*'); (*iii*) that can be the object of an inspiring event (as in '*Hamlet inspired many movies*').

Note that in our reverse engineering process we have discovered that popularity is a property that books can have, which is expressed as follows:

popularity ∈ book . **hasProp**

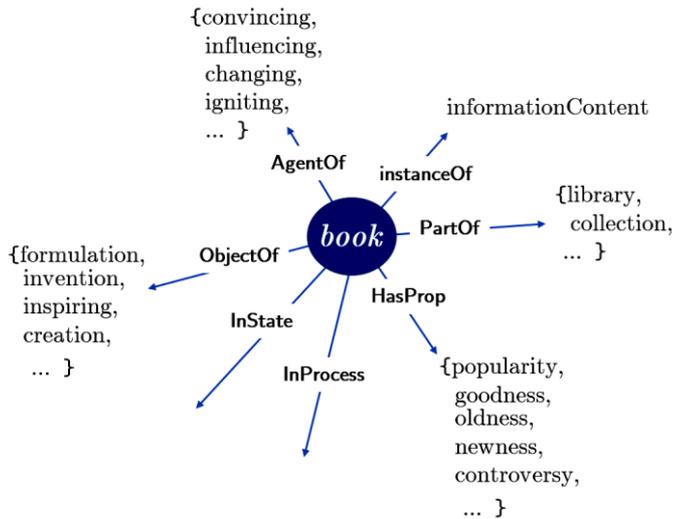

**Figure 3**. The primitive and linguistically agnostic relations as the dimensions of word meaning (in this case of one meaning of the word

We can use GPT-4 to generate some of these vectors along the various dimensions. The data in figure 4 is obtained by asking GPT-4 to provide 25 "plausible" (or "sensible") replacements for the [**MASK**] in the given sentences along three dimensions, namely **AgnetOf**, **ObjectOf** and **HasProp**, respectively.

Note that using this strategy we can also 'discover' the underlying ontology that seems to be implicit underneath our ordinary language. In figure 5 below we apply masking to generate the most plausible actions that a computer, a car, and a couch can be the object of. Note that while the three types of objects can be the objects of ASSEMBLE (we 'assemble' computers, cars, and couches), we can sensibly say a computer or a car is RUNNING (or that a computer or a car is OFF) but the same is not true of a couch.

| The book has [MASK] millions of people | Jon has [MASK] the book | Das Kapital was a very [MASK] book |
|---|---|---|
| 1. influenced | 1. wrote | 1. influential |
| 2. inspired | 2. criticized | 2. impactful |
| 3. educated | 3. endorsed | 3. controversial |
| 4. affected | 4. debated | 4. challenging |
| 5. engaged | 5. read | 5. analytical |
| 6. perplexed | 6. quoted | 6. detailed |
| 7. challenged | 7. discussed | 7. scholarly |
| 8. reached | 8. interpreted | 8. complex |
| 9. enlightened | 9. appreciated | 9. profound |
| 10. motivated | 10. translated | 10. radical |
| 11. stirred | 11. reviewed | 11. enlightening |
| 12. provoked | 12. studied | 12. dense |
| 13. intrigued | 13. analyzed | 13. thought-provoking |
| 14. alarmed | 14. examined | 14. significant |
| 15. shaped | 15. dismissed | 15. rigorous |
| 16. guided | 16. understood | 16. comprehensive |
| 17. fascinated | 17. refuted | 17. polemical |
| 18. informed | 18. praised | 18. controversial |
| 19. captivated | 19. digested | 19. transformative |
| 20. provoked | 20. researched | 20. critical |
| 21. challenged | 21. referenced | 21. pivotal |
| 22. transformed | 22. challenged | 22. historical |
| 23. touched | 23. summarized | 23. theoretical |
| 24. awakened | 24. defended | 24. intricate |
| 25. stimulated | 25. bought | 25. philosophical |

**Figure 4**. Querying GPT-4 to complete contexts with plausible actions/relations and properties that can plausibly (sensibly) be said of (or apply to) a book. Starting with the first column we can see that as an agent of some action or activity, a book can influence, inspire, motivate, educate, etc. people; as the object of some activity, a book can be translated, interpreted, examined, refuted, etc. and finally, a book can have the property (or can be described as being) significant, critical, historical, influential, controversial, etc.

This tells us that while a `computer` and a `car` must have some common supertype, these two types seem to eventually belong to a different branch from `couch` although the three objects must have a common supertype at some level of abstraction since they can all be 'assembled' (see figure 6). Analyzing the sets generated from this bottom-up reverse engineering would easily yield such an ontological structure.

| R(x, car) | R(x, computer) | R(x, couch) |
|---|---|---|
| 1. INNOVATE | 1. USE | 1. PURCHASE |
| 2. CONCEPTUALIZE | 2. OPERATE | 2. ASSEMBLE |
| 3. OPERATE | 3. PROGRAM | 3. MOVE |
| 4. UTILIZE | 4. BUILD | 4. UPHOLSTER |
| 5. SKETCH | 5. DEVELOP | 5. CLEAN |
| 6. DESIGN | 6. IMPROVE | 6. ADVERTISE |
| 7. BUILD | 7. ENGINEER | 7. SELL |
| 8. CRAFT | 8. CREATE | 8. POSITION |
| 9. CREATE | 9. INVENT | 9. ADMIRE |
| 10. MODIFY | 10. DESIGN | 10. INSTALL |
| 11. VISUALIZE | 11. MANUFACTURE | 11. INHERIT |
| 12. IMPROVE | 12. PRODUCE | 12. REPLACE |
| 13. DEVELOP | 13. INNOVATE | 13. COVER |
| 14. ENGINEER | 14. REFINE | 14. PHOTOGRAPH |
| 15. DRIVE | 15. EVOLVE | 15. REARRANGE |
| 16. RIDE | 16. ADVANCE | 16. INSPECT |
| 17. CONSTRUCT | 17. UTILIZE | 17. SIT ON |
| 18. ASSEMBLE | 18. MAINTAIN | 18. DISLIKE |
| 19. FABRICATE | 19. UPGRADE | 19. DONATE |
| 20. EVOLVE | 20. ENHANCE | 20. CHOOSE |
| 21. REFINE | 21. EXPAND | 21. FIND |
| 22. PIONEER | 22. OPTIMIZE | 22. DELIVER |
| 23. NAVIGATE | 23. ASSEMBLE | 23. STAIN |
| 24. PROTOTYPE | 24. STUDY | 24. VACUUM |
| 25. PLAN | 25. COMMERCIALIZE | 25. RESTORE |

**Figure 5**. Actions (verbs) that are sensible to say of a car, a computer and a couch.

To appreciate how the ontological structure implicit in ordinary language can be used in handling very complex phenomena in natural language we give one example here that involves what is called *metonymy*. Consider the following sentence (said by a waiter in some diner), and its commonsense interpretation:

*The loud omelet wants a beer.*
  → *The loud [**person eating the**] omelet wants a beer.*

How is it that speakers of ordinary language manage to uncover the missing but implicitly assumed (bold) text? This is done as follows:

1. the 'want' relation has the type constraint WANT(person, entity)
2. the text speaks of an omelet wanting a beer, i.e. WANT(omelet, beer)
3. WANT(omelet, beer) and WANT(human, entity) must now be *unified*
4. beer and entity unify since beer $\prec$ beverage $\prec$ ... $\prec$ entity
5. omelet and person do not unify since neither is a subtype of the other
6. the most salient relationship between person and omelet is retrieved
7. we have EAT(person, food) and omelet $\prec$ ... $\prec$ food
8. thus WANT(omelet, beer) $\rightarrow$ EAT(person, omelet) $\land$ WANT(person, beer)
9. *the 'loud omelet'* is now '*the loud person eating the omelet*'

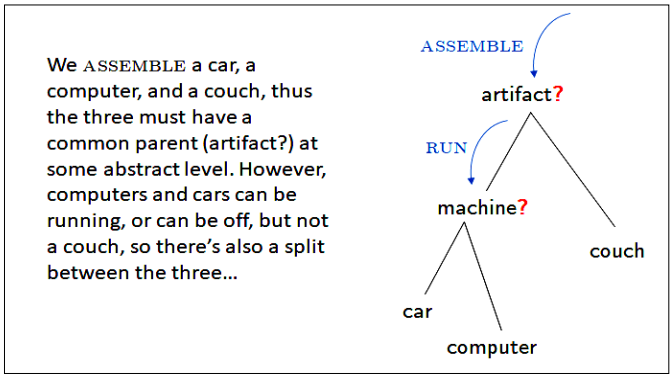

**Figure 6**. A computer, a car, and a couch can be assembled, so at some level of abstraction the three types must have a common parent (artifact?). However, cars and computers, although not couches, RUN and can be described by being OFF and so they eventually must be in different branches.

For a more detailed discussion on how the ontological structure implicit in ordinary language can be used in handling such complex phenomena in natural language the reader can consult (Saba, 2020).

## 6   Weights and Similarity

In figure 4 above we saw that a book can be described by the properties influential and profound, among others. That is,

influence ∈ book **. hasProp**
profoundness ∈ book **. hasProp**

Note, however, that one of these two properties might be more typical than another when it comes to books. That is, it might be the case that in our ordinary language use we speak of influential books more than we speak of profound books. That is, in general we might have

$(w_1,$ influence$) \in$ book **. hasProp**
$(w_2,$ profoundness$) \in$ book **. hasProp**

where $w_1 > w_2$ would indicate that influence is used when describing books more than profoundness – or that we say '*an influential book*' much more than we say '*a profound book*'. As such, the similarity of two concepts, $C_1$ and $C_1$ can be easily computed. First, consider two dimensions $C_1$ **.** $\mathbf{D}_1$ and $C_2$ **.** $\mathbf{D}_2$ for two concepts $C_1$ and $C_1$ (for example, book **. HasProp**$_1$ and magazine **. HasProp**$_2$). The sets along these dimensions look like this:

$C_1$ **.** $\mathbf{D}_1 = \{(w_{11}, p_{11}), (w_{12}, p_{12}), \ldots \}$
$C_2$ **.** $\mathbf{D}_1 = \{(w_{21}, p_{21}), (w_{22}, p_{22}), \ldots \}$

A feature set for the concepts $C_1$ and $C_2$ along the dimension $\mathbf{D}_1$ is then computed as follows:

FEATURESET($C_1, C_2, \mathbf{D}_1$)
$= C_1$ **.** $\mathbf{D}_1 \times C_1$ **.** $\mathbf{D}_1$
$= \{\langle(w_1, p_1), (w_2, p_2)\rangle \mid (p_1 = p_2) \wedge (w_1, p_1) \in C_1$ **.** $\mathbf{D}_1 \wedge (w_2, p_2) \in C_2$ **.** $\mathbf{D}_1\}$

For example, for one specific meaning of 'book' (see figure 3) and one specific meaning for 'publication' we could have the following:

*book*$_1$ **. HasProp** $= \{(0.75,$ popularity$), (0.73,$ controversy$), \ldots \}$
*publication*$_3$ **. HasProp** $= \{(0.72,$ popularity$), (0.71,$ controversy$), \ldots \}$

The feature set along the **HasProp** dimension is then:

FEATURESET($book_1$, $publication_3$, **HasProp**)
= {⟨(0.75, popularity), (0.72, popularity)⟩,
  ⟨(0.73, controversy), (0.71, controversy)⟩,
  ...
  }

The similarity along the **HasProp** dimension can now be computed as follows:

DSIMILARITY$_{\text{HasProp}}$($book_1$, $publication_3$)
= **sum**([FSIMILARITY($p$) for p in $fs$]) / |$fs$|
  where
  fs = *featureSet*($book_1$, $publication_3$, **HasProp**)
  FSIMILARITY(⟨($w_1$, $p_1$), ($w_2$, $p_2$)⟩) = **if** $p_1$ == $p_2$ **then** $1 - $ **abs**($w_1 - w_2$)
                                           = **else** 0

The final similarity between $book_1$ and $publication_3$ is then a weighted average of the similarity across all dimensions:

CSIMILARITY($book_1$, $publication_3$)
= **sum**([DSIMILARITY$_\mathbf{D}$($book_1$, $publication_3$) for **D** in *dims*]) / |*dims*|
  where
  *dims* = {**HasProp**, **AgentOf**, **ObjectOf**, **InState**, **PartOf**, ... }

One final note regarding concept similarity is that the above similarity is based on linguistic dimensions – that is, it is a similarity based on how we sensibly speak about concepts in our ordinary language. Thus, it will be expected that a book and a publication, for example, are quite similar, since almost anything that can sensibly be said of a book can also be said of a publication, and vice versa. However, this says nothing about (specific) instances of these objects. That is, linguistic features tell us that *influential* can sensibly be said of both a book and a publication but says nothing about that truth of a specific statement, such as '*Das Kapital is an influential book*'. In other words, while it is sensible to say *influential* of *Das Kapital*, whether or not that is a true statement is another issue. Thus the similarity we have discussed thus far is a concept (or type) similarity and not an instance (or object) similarity, which is another issue altogether the details of which we have to leave for another time.

## 7   Concluding Remarks

Large language models (LLMs) have proven that a bottom-up reverse engineering of language at scale is a viable approach. However, due to their subsymbolic nature, LLMs do not provide us with an explainable model of how language works nor how we externalize the thoughts we contemplate in language. The idea of a bottom-up reverse engineering of language, which LLMs proved to be viable approach could however be done in a symbolic setting, as has been suggested previously going back to Frege. The obvious and ideal solution, therefore, would be to combine the advantages of a bottom-up reverse engineering approach with an explainable symbolic representation, as we have done in this paper. How the symbolic dimensions of word meanings we discussed in this paper are used in language understanding would be the subject of future work.